\documentclass[letterpaper,10pt,conference]{ieeeconf}

\IEEEoverridecommandlockouts
\usepackage{cite}
\usepackage{graphicx}
\usepackage{multirow}      
\usepackage{booktabs}      
\usepackage{array}
\usepackage{booktabs}
\usepackage{amssymb}
\usepackage{subcaption}
\usepackage{caption}
\usepackage{tabularx} 
\usepackage{amsmath}
\usepackage{makecell}
\usepackage{wrapfig}
\usepackage{balance}

\title{\LARGE \bf Knowing When to Stop: Adaptive Action Chunking via Internal Cross-Attention Dynamics in VLAs}

\author{
  Runze Xu$^{1}$, Xiaolong Shan$^{1}$, Shuang Dai$^{1}$,
Yu Wang$^{1}$, and Jincheng Yu$^{1}$%
\thanks{$^{1}$Tsinghua University}%
}

\begin{document}
\maketitle
\thispagestyle{empty}
\pagestyle{empty}


\begin{abstract}
Action chunking is a standard execution strategy in modern Vision-Language-Action (VLA) frameworks, but fixed execution horizons impose a trade-off between efficiency and accuracy. Short chunks require frequent inference and may cause oscillatory behavior, whereas long chunks can become misaligned with newly observed states. We address this limitation with an adaptive action chunking approach based on internal cross-attention dynamics in the action expert. We observe that, as the prediction horizon extends, action-to-observation cross-attention becomes increasingly dispersed and its entropy rises toward a plateau. This pattern is associated with higher action prediction error and provides an online signal that the current observation offers limited grounding for further open-loop execution. Based on this observation, we introduce a training-free truncation mechanism that detects sustained high-entropy plateaus and dynamically selects the execution horizon during inference. The method uses attention weights already computed by the policy and introduces negligible additional overhead. Evaluations on $\pi_{0.5}$ and X-VLA across RoboTwin 2.0, LIBERO, and three real-world manipulation tasks show improved average task success over fixed-horizon and adaptive chunking baselines, while preserving efficient closed-loop control. These results show that cross-attention dynamics can provide a practical internal signal for adaptive action execution in VLAs.
\end{abstract}


\section{Introduction}

Vision-Language-Action (VLA) models have demonstrated significant success in robotic manipulation by integrating large-scale multimodal pre-training with physical execution\cite{black2024pi0,black2025pi05,liu2024rdt,liu2026rdt2,nvidia2025gr00tn1openfoundation,bu2025agibot,bu2025univla,team2025gemini,shukor2025smolvla}. These models typically combine a Vision-Language Model (VLM) backbone for high-level reasoning with a specialized action expert for fine-grained control. To capture multimodal trajectory distributions, action experts often employ generative denoising techniques such as diffusion\cite{sohl2015deep,ho2020denoising,song2020denoising} or flow matching\cite{lipman2024flow,lipman2022flow}. Conditioned on VLM hidden states, these experts sample continuous action sequences representing future trajectories rather than isolated points.

\begin{figure}[t]
  \centering
  \includegraphics[width=\linewidth]{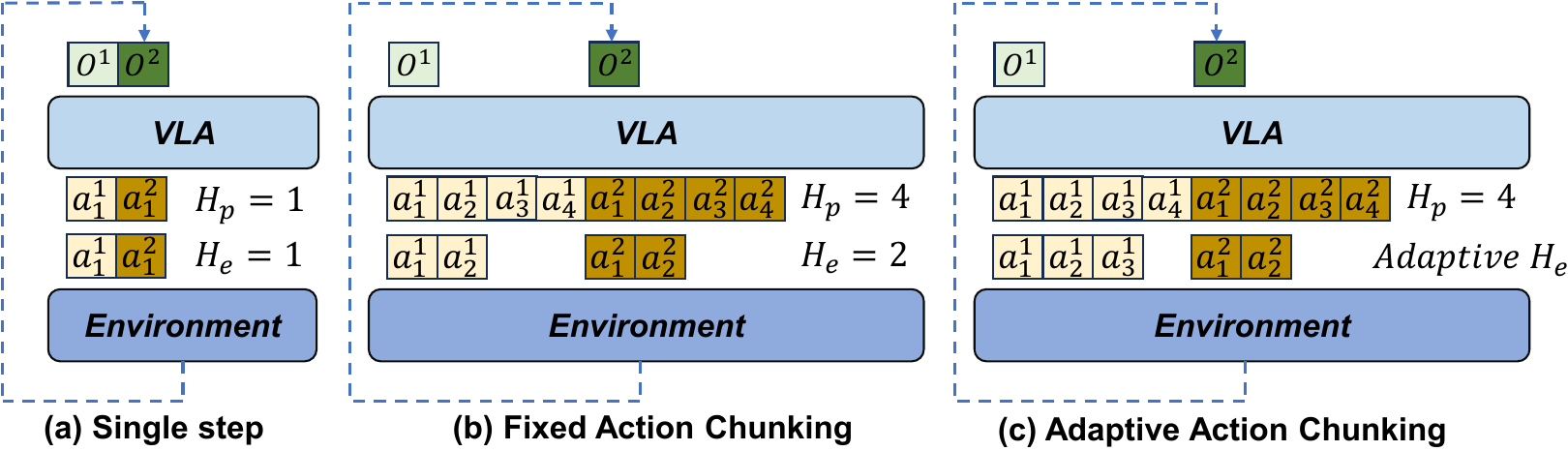}
  \caption{\textbf{Different action execution paradigms.} The model receives observation $O^t$ and predicts a sequence of actions $\{a^t_{j}\}$ over a predicted horizon $H_p$. (a) Single step: Predicts and executes one action per observation. (b) Fixed Action Chunking: Executes a fixed number of actions from the predicted horizon. (c) Adaptive Action Chunking: Dynamically adjusts the execution horizon to balance reactivity and efficiency.}
  \label{fig:background}
\vspace{-10pt}
\end{figure}

This serialized output paradigm effectively utilizes action chunking, a concept introduced by ALOHA \cite{zhao2023learning} to address the challenges of error accumulation and temporal interference in imitation learning. In single-step regimes (Fig.~\ref{fig:background}a), minor deviations in predicted actions accumulate over successive steps, leading to large divergence. Additionally, single-step policies often fail to capture temporal logic within training trajectories such as varying rhythms. In contrast, using action chunks (Fig.~\ref{fig:background}b) as the primary units of prediction and execution suppresses error propagation by reducing replanning frequency and improves the modeling of complex temporal patterns. During inference, the model generates an action sequence over a temporal window $H_p$, termed the predicted horizon. To maintain reactivity to environmental changes, the robot typically executes only the first $H_e$ steps, defined as the execution horizon\cite{chi2025diffusion}.

However, fixed execution horizons necessitate a trade-off between precision and efficiency \cite{so2026improving,liang2026adaptive,wang2026vla,liu2025bidirectional}. Short horizons increase computational costs through frequent inference and may induce oscillatory behaviors. Conversely, excessively long horizons misalign execution with real-time observations; as visual feedback relevance decays, predictive accuracy collapses \cite{black2026real,liu2026learning}. Furthermore, optimal horizons vary both across tasks and within single-task stages, such as transitioning from rapid approach to fine-grained manipulation \cite{liang2026adaptive}. Current reliance on manual tuning or static heuristics \cite{black2025pi05,nvidia2025gr00tn1openfoundation} thus limits VLA generalization in diverse scenarios.

\begin{figure*}[t]
  \centering
  \includegraphics[width=\linewidth]{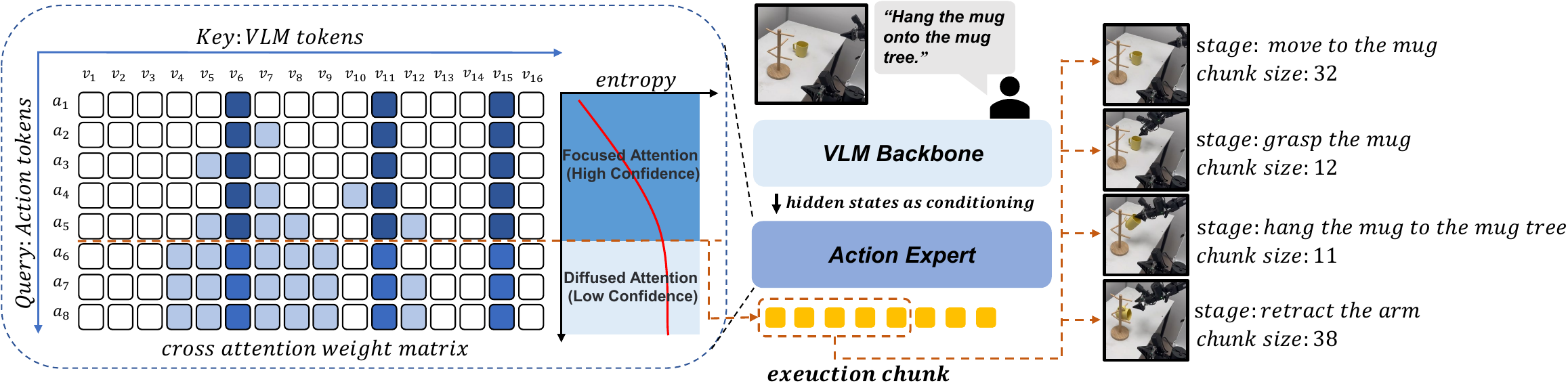}
  \caption{\textbf{Method Overview.} Our framework dynamically determines action chunk sizes by monitoring the cross-attention entropy within the action expert. Conditioned on VLM states, the action expert identifies chunk boundaries by monitoring attention focus. This enables adaptive execution across task stages.}
  \label{fig:overview}
\vspace{-10pt}
\end{figure*}

To address the limitations of fixed horizons, we examine the internal attention allocation of action experts (Fig.~\ref{fig:overview}), focusing on cross-attention from action queries to VLM tokens. Across the predicted horizon, we observe a systematic change in this distribution. Early action tokens attend sparsely to a small set of perceptual and linguistic features, whereas later tokens distribute their attention over a broader context. The corresponding entropy typically rises with the action index and may eventually enter a high-value saturation regime. Empirically, sustained high entropy indicates reduced concentration on the current perceptual context and is associated with elevated offline action error. It therefore provides a practical proxy for identifying predictions that are at greater risk of becoming weakly grounded in the current observation. Based on this insight, we propose an adaptive truncation mechanism. This mechanism monitors cross-attention entropy during inference and dynamically determines the execution horizon $H_e$ when the entropy reaches its plateau. This approach establishes a dynamic closed loop between perception and action with near-zero computational overhead.

The primary contributions of this work are three-fold. (1) We characterize the horizon-wise dispersion of action-to-VLM attention in flow-based VLA policies and show that sustained high-entropy saturation is associated with increased offline action error, supporting its use as an empirical indicator of reduced perceptual grounding and elevated prediction risk. (2) Based on these findings, we design a training-free dynamic truncation mechanism that adaptively adjusts the execution horizon during inference without requiring architectural modifications. (3) Extensive evaluations across multiple state-of-the-art VLA models in both simulated and real-world environments demonstrate that our method achieves higher average success rates than the evaluated fixed-horizon baselines and adaptive baselines, while incurring little additional inference latency.


\section{Related Work}
\subsection{Vision-Language-Action (VLA) Models}
VLA architectures combine visual observations, language instructions, and robot actions within a shared policy framework. A common design uses a pre-trained VLM to encode semantic and perceptual context, which then conditions a dedicated action-generation module. Early systems formulate control as autoregressive prediction over discretized action tokens\cite{zitkovich2023rt,kim2024openvla}. This formulation directly reuses language-model training objectives, but action discretization and sequential decoding can limit precision and increase the cost of modeling long trajectories. More recent approaches employ continuous action experts based on diffusion or flow matching to represent multimodal action distributions and generate an entire future trajectory in parallel\cite{black2024pi0,black2025pi05,liu2024rdt,nvidia2025gr00tn1openfoundation,zheng2025x,zhao2025cot}. Their architectures range from decoupled designs, in which action queries attend to fixed VLM features, to coupled designs that jointly update VLM and action tokens through a shared attention stack.

Existing work has primarily focused on improving policy training and action generation, while typically relying on a manually selected, fixed execution horizon at inference time. In contrast, we study when predicted future actions cease to be sufficiently grounded in the current observation. Our method treats sustained dispersion in action-to-VLM attention as an empirical signal and uses it to adapt the action-execution horizon at inference time.
\begin{figure*}[h]
  \centering
  \begin{minipage}[c]{0.45\textwidth}
    \centering
    \subcaptionbox{$\pi_{0.5}$ attention weights across all layers at denoising step 10.\label{fig:cross_all}}{%
      \includegraphics[width=\linewidth]{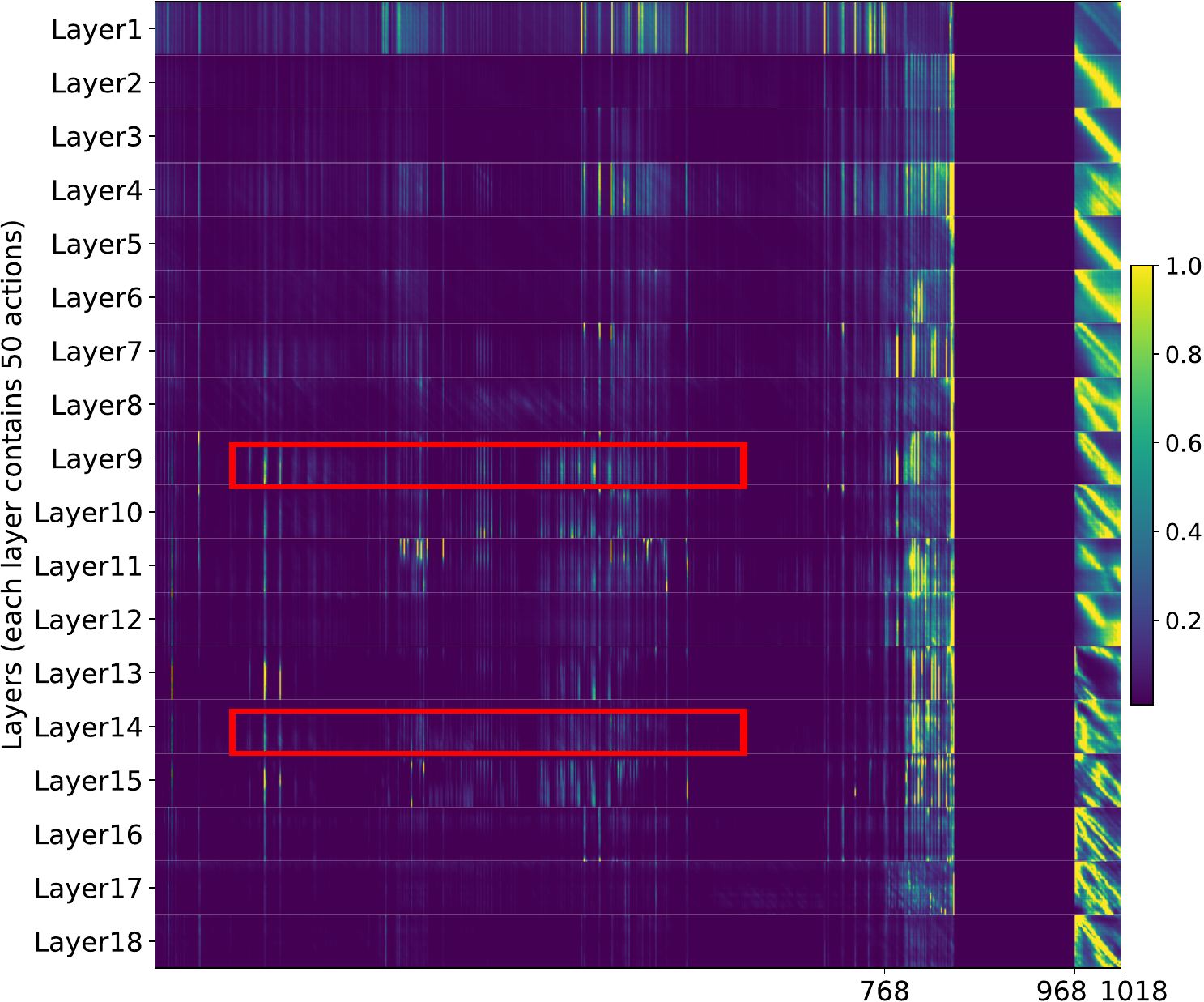}%
    }
  \end{minipage}
  \hfill
  \begin{minipage}[c]{0.53\textwidth}
    \centering
    \subcaptionbox{Layer 9.\label{fig:attn_layer9}}{%
      \includegraphics[width=\linewidth]{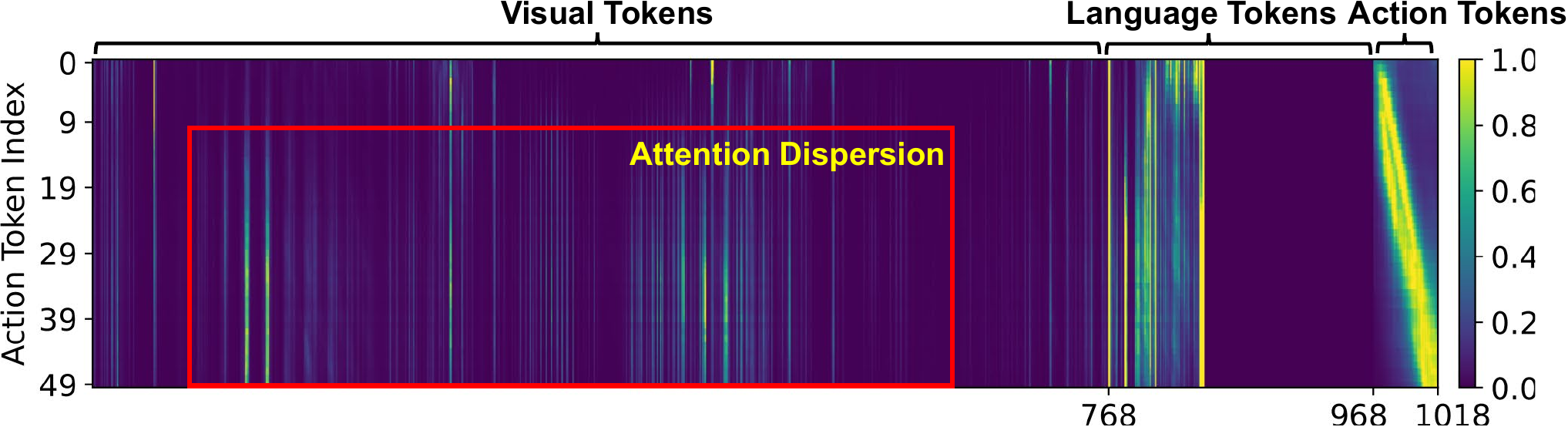}%
    }
    \vspace{0.1cm}
    \subcaptionbox{Layer 14.\label{fig:attn_layer14}}{%
      \includegraphics[width=\linewidth]{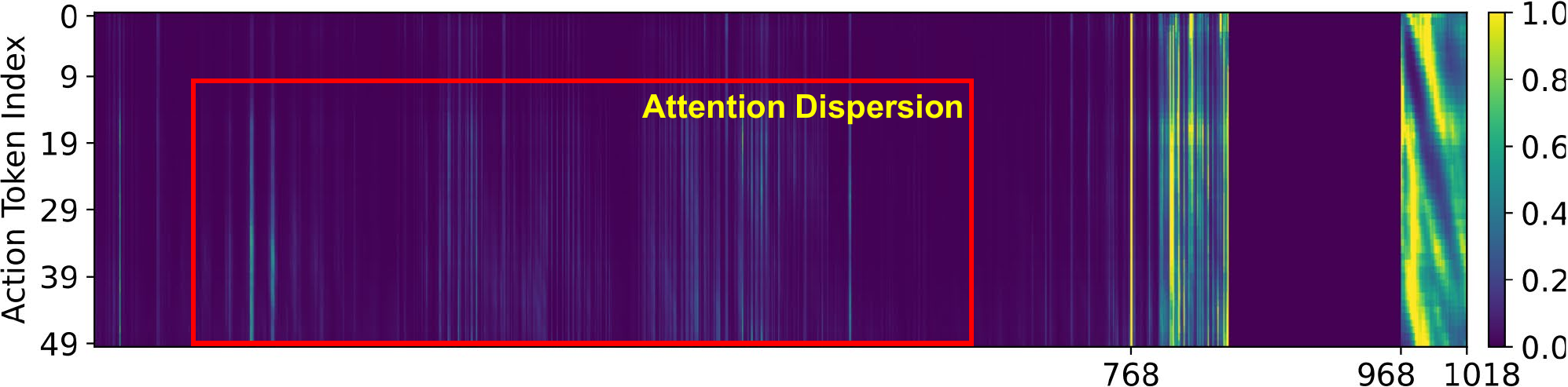}%
    }
  \end{minipage}
  \caption{\textbf{Visualization of attention dynamics within the action expert of $\pi_{0.5}$.} (a) provides a comprehensive view across all 18 layers, where the vertical axis represents 50 action queries per layer and the horizontal axis comprises 1,018 key tokens (768 visual, 200 language, and 50 action tokens). (b) and (c) present magnified views of Layer 9 and Layer 14. Red boxes highlight the pronounced attention dispersion for later-stage actions, whereas tokens with smaller indices exhibit negligible reliance on these visual features.}
  \label{fig:combined_attn}
\end{figure*}
\subsection{Action Chunking}
Unlike single-step action inference, action chunking\cite{lai2022action} generates and executes a sequence of actions in a single pass, which mitigates cumulative errors and enhances the temporal modeling capabilities of the model. Following its introduction in ALOHA\cite{zhao2023learning}, this technique became the standard paradigm for VLA models\cite{black2024pi0,black2025pi05,nvidia2025gr00tn1openfoundation,zheng2025x}. However, fixed execution horizons require tedious manual tuning for specific tasks and restrict adaptive generalization. While BID\cite{liu2025bidirectional} attempts to identify optimal chunk lengths, it maintains static horizons during execution, remaining unresponsive to task dynamics. \cite{liang2026adaptive} performs \textbf{multiple sampling (MS)} during the flow matching denoising process to estimate action uncertainty via entropy for dynamic truncation. Alternatively, \cite{wang2026vla} utilizes \textbf{self-attention (SA)} weights of action tokens relative to sequence anchors to evaluate confidence. This strategy essentially relies on internal sequence consistency but overlooks the root cause of action drift, specifically that current observations become stale for long-range predictions. To address these challenges, we propose an adaptive action chunking method based on the dispersion of cross-attention. By measuring the distribution characteristics of attention from action tokens to VLM tokens, we characterize prediction confidence with negligible computational burden. Experimental results demonstrate that our approach improves adaptive execution performance in multi-task environments without introducing substantial overhead.

\section{Methods}
\subsection{Unified Interaction Formulation in VLA}
To develop a general adaptive chunking mechanism, we first examine how different VLA architectures handle the interaction between actions and perceptions. While specific implementations vary, most modern frameworks rely on an attention-based interface to fuse multimodal information. For instance, $\pi_{0.5}$ adopts a decoupled paradigm where VLM tokens serve as static keys and values that remain unchanged across all layers. In contrast, X-VLA utilizes a coupled paradigm where VLM and action tokens are concatenated and updated jointly via self-attention.

Despite these architectural differences, the underlying action generation process can be unified as a cross-attentional retrieval. Specifically, as shown in Fig.~\ref{fig:overview}, the expert uses a set of \textbf{action queries} $\{a_j\}_{j=1}^{H_p}$ to extract information from \textbf{perceptual keys} $\{v_i\}_{i=1}^{N}$ provided by the VLM backbone. This interaction is captured by an attention weight matrix, where the row index $j$ represents the temporal step of the predicted action sequence, and the column index $i$ corresponds to the perceptual features of the observation. This unified formulation enables us to seek an internal signal within the attention matrix that is associated with reduced perceptual grounding of future actions. In the following sections, we instantiate our methodology using $\pi_{0.5}$ as the representative model to analyze cross-attention dynamics, with X-VLA-specific adaptations discussed in the experimental evaluation.

\subsection{Attention Dispersion and Entropy Dynamics}
\subsubsection{Qualitative Observation of Attention Maps}
We begin by visualizing the head-averaged action-query attention matrix
$\bar{\mathbf{A}}_l\in[0,1]^{H_p\times(N+H_p)}$
at each layer $l$ of the action expert. As illustrated in Fig.~\ref{fig:cross_all}, the vertical axis represents the $H_p=50$ action queries at each layer, while the horizontal axis comprises both VLM and action-token keys. Specifically, the first $N=968$ columns correspond to 768 visual tokens and 200 language tokens, and the remaining 50 columns correspond to action tokens. Because the qualitative patterns are consistent across denoising steps, we visualize the final denoising step (step 10). The cross-modal component analyzed by our method is the action-to-VLM submatrix $\mathbf{W}_l$, corresponding to the first $N$ columns of $\bar{\mathbf{A}}_l$.

A global inspection of the intermediate and deep layers, specifically from Layer 9 to Layer 15, reveals a systematic shift in attention focus. For actions at the beginning of the sequence, the attention exhibits sharp and concentrated saliency, primarily localized on language tokens. However, as the action index increases, the attention focus shifts toward visual tokens and becomes significantly dispersed. We magnify Layer 9 (Fig.~\ref{fig:attn_layer9}) and Layer 14 (Fig.~\ref{fig:attn_layer14}) as representative examples. As indicated by the red boxes, attention for later actions is distributed across nearly the entire visual field. This intra-layer transition shows an empirically observed regime in which later action tokens are less selectively grounded in the current perceptual context; we therefore examine attention dispersion as a candidate empirical risk signal associated with reduced perceptual grounding of future actions.
\subsubsection{Quantification via Cross-Attention Entropy}
To quantify this phenomenon, we isolate the action-to-VLM dependency from the full attention matrix. Let $\alpha_{l,j,i} = W_l[j, i]$ denote the attention weight of the $j$th action token on the $i$th VLM token $v_i$ at layer $l$. We introduce cross-attention weight entropy to quantify how focus is distributed across VLM tokens, where higher values indicate a more divergent allocation. To mitigate layer-specific noise and enhance stability, we aggregate the action-to-VLM attention coefficients along the layer axis. These values are subsequently normalized to obtain a probability distribution $p_{j,i}$ describing the relative attention assigned to each VLM token $v_i$ by action token $a_j$: 
\begin{equation}
  \label{eq:eq1}
p_{j, i} = \mathrm{Norm} \left( \sum_{l=1}^{L} \alpha_{l, j, i} \right), \quad i \in \{1, \dots, N\},
\end{equation}
where $\mathrm{Norm}(\cdot)$ ensures that $\sum_{i=1}^{N} p_{j, i} = 1$ for each action token $a_j$.
Based on this distribution, the cross-attention weight entropy $\mathcal{E}_{j}$ for each action token $a_j$ is computed as 
\begin{equation}
  \label{eq:eq2}
\mathcal{E}_{j} = - \sum_{i=1}^{N} p_{j, i} \ln p_{j, i}.
\end{equation}
Because $p_{j,i}$ is defined over $N$ retained VLM tokens, its entropy is bounded by $0\leq\mathcal{E}_j\leq\ln N$, with equality at the upper bound when attention is uniformly distributed. This architecture-dependent bound provides a normalized reference for distinguishing a genuinely dispersed high-entropy plateau from a locally flat but still concentrated attention pattern.

\begin{figure*}[t] 
  \centering
  \begin{subfigure}[b]{0.32\linewidth}
    \centering
    \includegraphics[height=0.115\textheight]{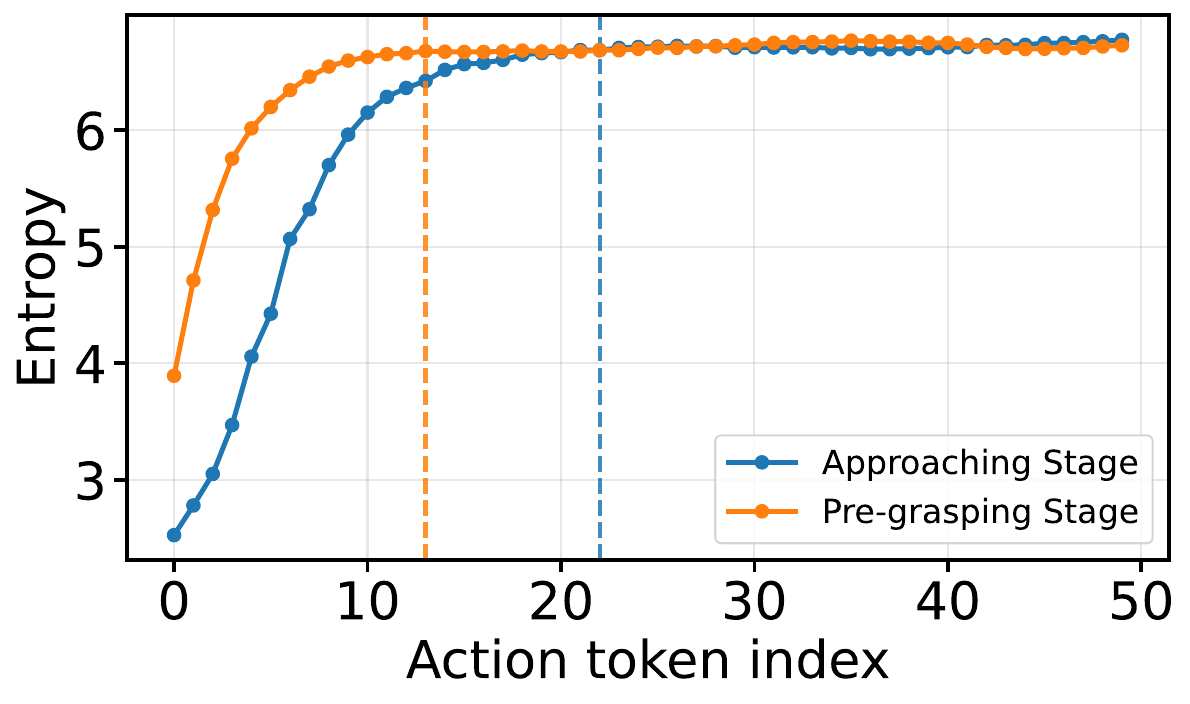}
    \caption{Stage-dependent entropy.}
    \label{fig:sub_stack_1}
  \end{subfigure}
  \hfill
  \begin{subfigure}[b]{0.26\linewidth}
    \centering
    \includegraphics[height=0.115\textheight]{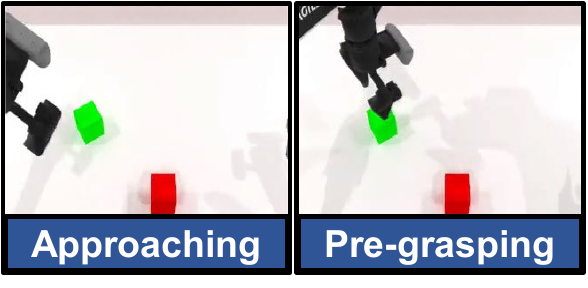}
    \caption{Different stages.}
    \label{fig:sub_stack_2}
  \end{subfigure}
  \hfill
  \begin{subfigure}[b]{0.4\linewidth}
    \centering
    \includegraphics[height=0.115\textheight]{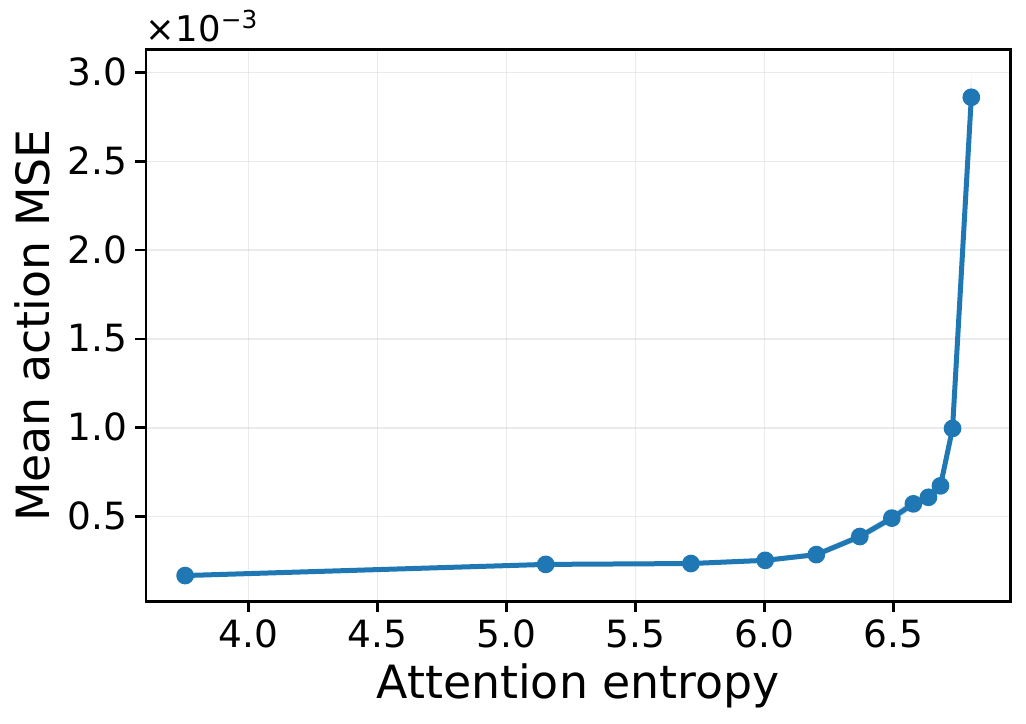}
    \caption{Entropy versus action MSE.}
    \label{fig:binned_entropy_mse}
  \end{subfigure}
  \caption{\textbf{Action entropy analysis.} (a) Attention entropy exhibits distinct growth rates across the task stages visualized in (b). Precise interactions (pre-grasping) lead to a sharper rise and earlier saturation than coarse movements (approaching). (c) Offline evaluation on the validation set shows that the action-prediction MSE rises sharply as cross-attention entropy approaches its theoretical maximum. Samples in the highest-entropy quartile exhibit 10.4 times the mean MSE and
12.1 times the MSE standard deviation of remaining samples.}
  \label{fig:entropy_index}
\end{figure*}

We analyze the evolution of cross-attention weight entropy $\mathcal{E}$ relative to the action index $j$. As illustrated in Fig.~\ref{fig:sub_stack_1}, entropy consistently increases with the action index before reaching a stable plateau. This plateau indicates an empirically observed regime in which action-to-VLM attention is highly dispersed and temporally stable. We restrict this interpretation to the observed attention pattern and evaluate its association with action error below. The onset of this regime varies with the task stage. During coarse approach maneuvers, entropy increases gradually and attention remains concentrated over a longer action prefix. In contrast, during precise pre-interaction phases, entropy rises sharply and plateaus at a much earlier index. These observations motivate a stage-dependent execution rule: a fixed horizon cannot accommodate changes in the action-to-observation grounding pattern across different manipulation stages.
\subsubsection{Entropy as Accuracy Proxy}
To test whether entropy is associated with action prediction error rather than merely action index position, we conduct an inference analysis on held-out demonstration trajectories from the validation set. At each evaluated frame, the policy predicts a horizon of 50 actions. For every valid horizon index, we compare the predicted action with the ground-truth action at the corresponding future time step in the recorded trajectory and associate the resulting MSE with that action token's cross-attention entropy, with predictions extending beyond the end of a trajectory excluded. Entropy and MSE exhibit significant positive dependence (Pearson $r=0.187$, Spearman $\rho=0.432$; both $p<10^{-120}$). After regressing both variables on the within-chunk action index, their partial correlation remains significant ($r=0.214$), showing that the relationship is not explained solely by later actions being harder to predict.

To construct Fig.~\ref{fig:binned_entropy_mse}, we sort all valid pairs by entropy and partition them into 12 equal-frequency bins. Each point reports the mean entropy and mean MSE of all pairs within one bin, thereby showing how average action prediction error varies across different entropy ranges while keeping the number of samples in each range comparable. The resulting relationship is strongly nonlinear: the average MSE rises sharply in the high-entropy regime near $\mathcal{E}=6.5$, which is approximately $95\%$ of the theoretical entropy upper bound $\ln(968)$. Moreover, samples in the highest-entropy quartile exhibit 10.4 times the mean MSE and 12.1 times the MSE standard deviation of samples in the remaining three quartiles. This marked increase in both prediction error and error variability indicates that high cross-attention entropy is associated with elevated prediction risk, supporting its use as an online signal for adaptive truncation.

\subsection{Adaptive Action Chunking}
Based on the observed entropy dynamics, we propose an Adaptive Action Chunking strategy that dynamically selects the execution horizon during inference. Given the cross-attention entropy sequence
$\mathbf{E}=\{\mathcal{E}_1,\mathcal{E}_2,\ldots,\mathcal{E}_{H_p}\}$
computed at the final denoising step, the method identifies the earliest point at which attention enters a sustained high-dispersion regime.

To reduce noise in the entropy sequence, we first apply a moving-average
filter with a window size of $k$:
\begin{equation}
\bar{\mathcal{E}}_j =
\frac{1}{k}\sum_{m=0}^{k-1}\mathcal{E}_{j+m},
\qquad
j \in \{1,\ldots,H_p-k+1\}.
\label{eq:entropy_smoothing}
\end{equation}

We then compute the first-order change of the smoothed entropy sequence:
\begin{equation}
\Delta\bar{\mathcal{E}}_j
=\bar{\mathcal{E}}_{j+1}-\bar{\mathcal{E}}_j.
\label{eq:entropy_difference}
\end{equation}

A low entropy gradient alone is not sufficient to indicate saturation,
because entropy can also vary slowly while remaining concentrated.
Therefore, we detect a plateau only when the smoothed entropy is both
high and temporally stable. Specifically, we define the set of candidate
plateau locations as
\begin{equation}
\begin{aligned}
\mathcal{C} =
\Bigl\{
j \in \{1,\ldots,H_p-2k+1\}
\ \Bigm|\ 
& \min_{0\leq n<k}
\bar{\mathcal{E}}_{j+n}
\geq \eta\ln N, \\
& \frac{1}{k-1}
\sum_{n=0}^{k-2}
\left|\Delta\bar{\mathcal{E}}_{j+n}\right|
< \tau
\Bigr\}.
\end{aligned}
\label{eq:plateau_candidates}
\end{equation}
where $\eta$ is the normalized high-entropy threshold and $\tau$ is the
stability threshold. The first condition requires every smoothed entropy
value in the $k$-step window to exceed a fraction $\eta$ of the maximum
entropy $\ln N$. The second condition requires the average absolute entropy
change within the same window to remain small. The adaptive execution horizon is then defined as
\begin{equation}
H_e =
\begin{cases}
\min(\mathcal{C}) + k, & \mathcal{C}\neq\varnothing,\\
H_p, & \mathcal{C}=\varnothing.
\end{cases}
\label{eq:adaptive_horizon}
\end{equation}
The offset $+k$ accounts for the $k$-step window used to detect the
plateau. Consequently, a low-entropy plateau does not trigger
truncation: if no window satisfies both conditions, the policy executes
the full predicted horizon $H_p$.

By selecting the execution horizon from the observed action-to-VLM
attention dynamics, the method retains longer action chunks while
attention remains concentrated and truncates only after detecting a
sustained high-dispersion regime associated with increased prediction
error. The rule uses attention weights already produced by the action
expert and requires neither retraining nor additional sampling.

\begin{figure}[t]
  \centering
  \includegraphics[width=\linewidth]{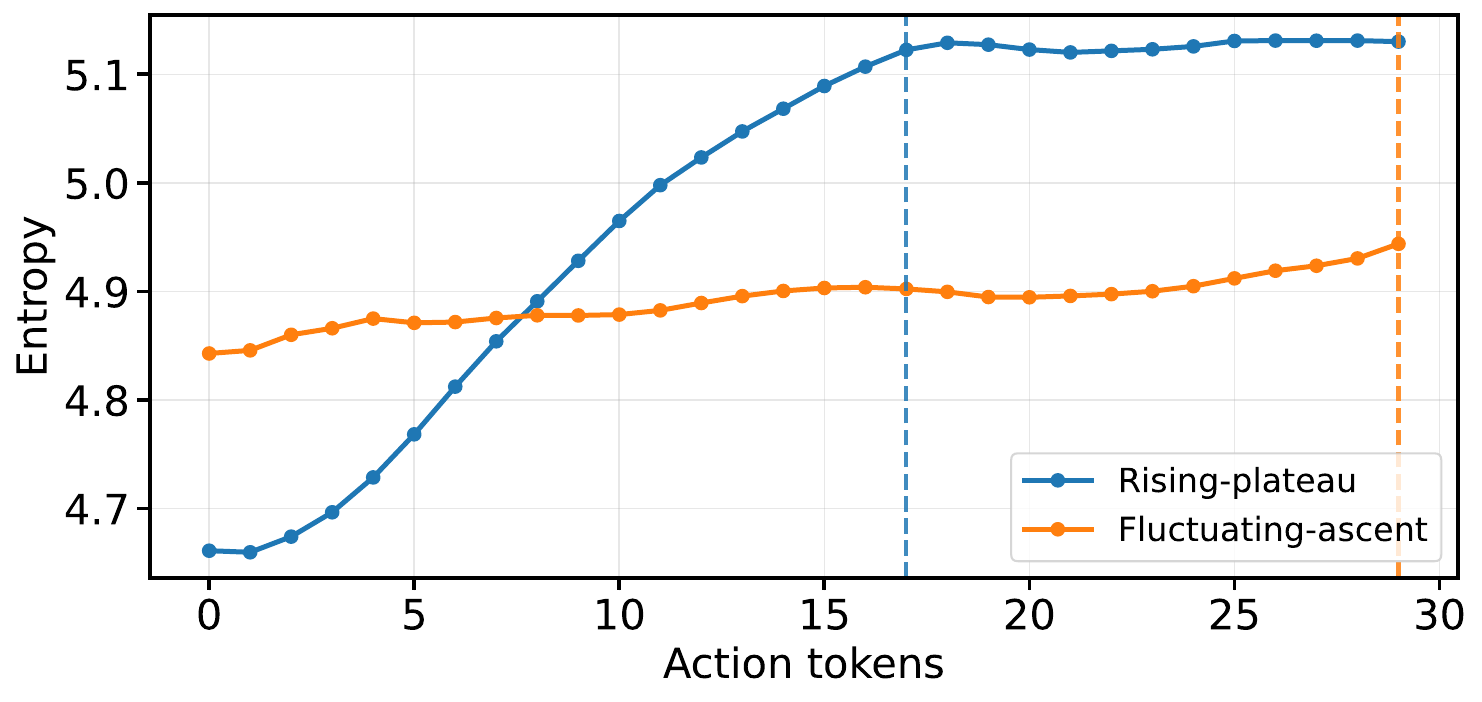}
  \caption{\textbf{Entropy patterns in X-VLA.} Two distinct patterns are observed: the rising-plateau pattern reaches a sustained high-entropy regime, whereas the fluctuating-ascent pattern continues to vary without forming a stable plateau.}
  \label{fig:entropy_xvla}
\end{figure}

\begin{figure}[t]
  \centering
  \includegraphics[width=\linewidth]{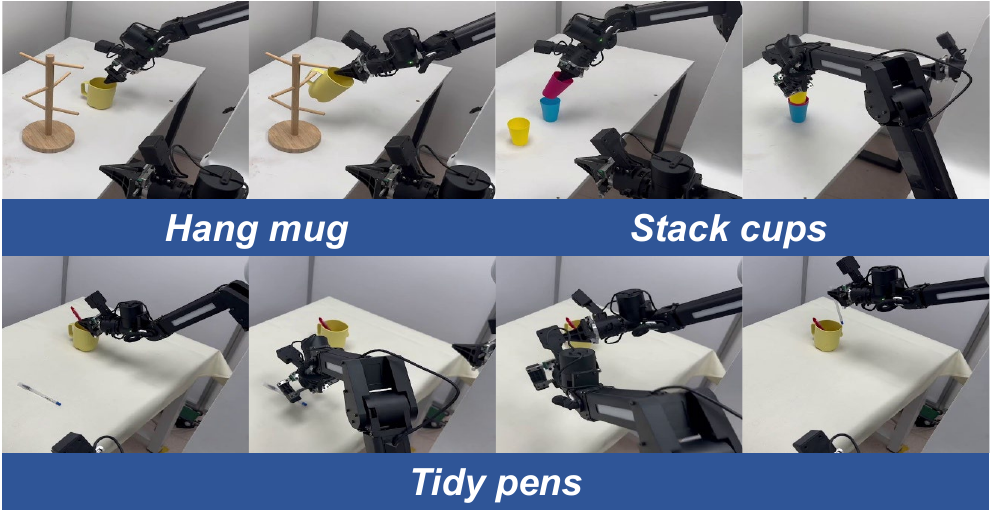}
  \caption{\textbf{Real-world tasks.} We evaluate the generalizability of our adaptive chunking method across three representative real-world tasks: Hang mug, Stack cups, and Tidy pens.}
  \label{fig:real_tasks_small}
\end{figure}

\begin{table}[t]
\centering
\caption{RoboTwin 2.0 task statistics.}
\label{tab:task_statistics}
\small
\renewcommand{\arraystretch}{1.0}
\begin{tabular*}{\linewidth}{@{\extracolsep{\fill}}lcc}
\toprule
\textbf{Task} & \textbf{Avg. Frames} & \textbf{Horizon} \\
\midrule
Place phone stand    & 127.74 & Short \\
Place fan            & 147.66 & Short \\
Move can pot         & 151.68 & Short \\
Place dual shoes     & 227.94 & Medium \\
Place bread basket   & 238.80 & Medium \\
Hand over block      & 284.98 & Medium \\
Stack blocks two     & 313.58 & Medium \\
Blocks rank rgb      & 461.00 & Long \\
Stack blocks three   & 473.16 & Long \\
Put bottles dustbin  & 622.42 & Long \\
\bottomrule
\end{tabular*}
\end{table}

\begin{table*}[t]
\centering
\caption{Success Rate (\%) and Average Action Length for $\pi_{0.5}$ in RoboTwin 2.0 benchmark.}
\label{tab:results_pi05}
\small
\renewcommand{\arraystretch}{1.0}
\begin{tabular*}{\linewidth}{@{\extracolsep{\fill}}l ccccccc c}
\toprule
 & \multicolumn{7}{c}{\textbf{Success Rate (\%) by Execution Chunk}} & \textbf{Ours} \\
\cmidrule(lr){2-8}
\textbf{Task Name} & \textbf{10} & \textbf{15} & \textbf{20} & \textbf{25} & \textbf{SA} & \textbf{MS} & \textbf{Ours} & \textbf{Avg. Len.} \\
\midrule
Blocks rank rgb     & 80.0 & 62.0 & 52.7 & 43.3 & 58.3 & 74.7 & 77.3 & 20.83 \\
Put bottles dustbin & 89.3 & 90.0 & 84.7 & 82.7 & 87.3 & 78.0 & 91.3 & 19.54 \\
Stack blocks three  & 63.0 & 43.3 & 28.7 & 36.7 & 53.0 & 66.0 & 62.3 & 18.30 \\
Stack blocks two    & 91.7 & 85.0 & 82.0 & 77.3 & 85.0 & 81.3 & 89.3 & 20.31 \\
Hand over block     & 19.0 & 27.0 & 29.0 & 25.3 & 26.0 & 10.0 & 25.3 & 15.91 \\
Place dual shoes    & 31.0 & 37.0 & 37.0 & 38.7 & 36.0 & 33.7 & 39.3 & 17.10 \\
Place bread basket  & 55.7 & 54.7 & 55.7 & 55.0 & 52.3 & 48.0 & 57.3 & 14.88 \\
Move can pot        & 90.7 & 87.0 & 73.0 & 68.3 & 75.3 & 84.7 & 87.3 & 17.89 \\
Place fan           & 44.0 & 42.7 & 37.3 & 39.0 & 41.3 & 51.0 & 42.7 & 21.42 \\
Place phone stand   & 52.0 & 51.0 & 58.0 & 55.3 & 61.0 & 42.7 & 56.7 & 14.88 \\
\midrule
\textbf{Average}    & 61.6 & 58.0 & 53.8 & 52.2 & 57.6 & 57.0 & \textbf{62.9} & 18.11 \\
\bottomrule
\end{tabular*}
\end{table*}

\section{Experiments}
We evaluate our adaptive action chunking method using state-of-the-art $\pi_{0.5}$ and X-VLA models across the RoboTwin 2.0\cite{chen2025robotwin,mu2025robotwin} and LIBERO\cite{liu2023libero} benchmarks, as well as three real-world tasks. Systematic comparisons against fixed-length, multi-sampling\cite{liang2026adaptive} (MS), and self-attention-based\cite{wang2026vla} (SA) baselines demonstrate that our method improves success rates and generalizability.

\noindent\textit{Hyperparameters.} We use a common configuration of $k=5$, $\tau=0.01$, and $\eta=0.95$ across models, benchmarks, and tasks. The normalized high-entropy ratio $\eta$ is selected once according to the entropy--MSE relationship in Fig.~\ref{fig:binned_entropy_mse}, which shows a rapid increase in action error in the high-entropy regime, and is not tuned for individual tasks or architectures. The sole exception is $\pi_{0.5}$ on LIBERO: because this policy predicts only $H_p=10$ actions per query, we use a proportionally shorter smoothing window of $k=3$. All four LIBERO suites share this setting, while $\tau$ and $\eta$ remain unchanged.
\subsection{Simulation}
\subsubsection{RoboTwin 2.0 Experiments}
To ensure comprehensive assessment, we evaluate 10 representative RoboTwin 2.0 tasks, categorized by average episode length into short (3 tasks), medium (4 tasks), and long-horizon (3 tasks) tiers. The selected tasks and their average episode lengths are summarized in Tab.~\ref{tab:task_statistics}. For $\pi_{0.5}$, we fine-tune individual checkpoints for 40,000 steps (batch size 32) using 50 Aloha Agilex demonstrations per task. For X-VLA, we evaluate the open-source checkpoint trained on all 50 RoboTwin tasks. All evaluations are conducted in RoboTwin's clean mode using three random seeds per configuration, with 100 rollouts each.

\noindent\textit{Evaluation on $\pi_{0.5}$.}
The fine-tuned $\pi_{0.5}$ model generates a candidate sequence of $H_p = 50$ actions per inference step. We evaluate several fixed execution chunk lengths, specifically 10, 15, 20, and 25, while omitting longer horizons due to their significantly degraded success rates. Dynamic baselines include the multi-sampling method, which utilizes 20 iterations to estimate action entropy, and the self-attention-based approach. The mean success rates across three seeds are summarized in Tab.~\ref{tab:results_pi05}. Our method achieves a leading average success rate of 62.9\%, outperforming all fixed-chunk baselines and established methods such as self-attention-based (SA) and multi-sampling (MS). While fixed-chunk baselines can achieve competitive performance on individual tasks, their adaptability across a diverse multi-task suite is limited compared to our approach. Our method achieves the highest average success rate across the evaluated task suite, demonstrating both stability and effective generalization. It outperforms the strongest baseline (chunk size = 10) despite having a much longer average execution horizon of 18.11 steps. This indicates that stage-aware dynamic adjustment can improve success rates while reducing inference frequency. By extending execution during high-confidence phases, our strategy lowers computational overhead, providing a better trade-off between task precision and inference efficiency.

\noindent\textit{Evaluation on X-VLA.}
To further validate the generalizability of our method, we conduct the same experiments on X-VLA. X-VLA generates a candidate sequence of $H_p = 30$ actions at each inference step. We evaluate fixed execution horizons of 15, 20, 25 and 30, while omitting shorter horizons due to their low success rates. Although X-VLA employs coupled self-attention over concatenated VLM and action tokens, we apply the same extraction procedure described in Sec.~III. Specifically, we first average the action-query attention weights across heads to obtain $\bar{\mathbf{A}}_l$ and then extract the action-query-to-VLM-key block $\mathbf{W}_l$. For X-VLA, this block contains the attention assigned by each action query to the $N=200$ VLM tokens. We compute the normalized distribution $p_{j,i}$ and its entropy $E_j$ following Eq.~\ref{eq:eq1} and Eq.~\ref{eq:eq2}. As shown in Fig.~\ref{fig:entropy_xvla}, X-VLA exhibits both rising-plateau and fluctuating-ascent patterns. We apply exactly the same normalized criterion in Eq.~\ref{eq:plateau_candidates} as for $\pi_{0.5}$, without an architecture-specific entropy-increment threshold. A rising or fluctuating sequence that never forms a sustained high-entropy plateau leaves $\mathcal{C}$ empty and therefore executes the full 30-action horizon; only a sustained high-entropy plateau triggers truncation. Results in Tab.~\ref{tab:results_xvla} show that although fixed execution lengths achieve peak performance on individual tasks, their performance varies substantially across the task suite. Our method attains the highest average success rate without model-specific threshold design.

\begin{table}[t]
\centering
\caption{Success Rate (\%) for X-VLA.}
\label{tab:results_xvla}
\small
\renewcommand{\arraystretch}{1.0}
\setlength{\tabcolsep}{3pt}
\begin{tabular*}{\linewidth}{@{\extracolsep{\fill}}lccccc}
\toprule
 & \multicolumn{5}{c}{\textbf{SR (\%) by Execution Chunk}} \\
\cmidrule(lr){2-6}
\textbf{Task Name} & \textbf{15} & \textbf{20} & \textbf{25} & \textbf{30} & \textbf{Ours} \\
\midrule
Blocks rank rgb     & 33.0 & 84.0 & 85.7 & 82.7 & 84.0 \\
Put bottles dustbin & 46.0 & 72.7 & 80.0 & 80.0 & 71.3 \\
Stack blocks three  & 22.0 & 52.7 & 58.0 & 45.3 & 59.3 \\
Stack blocks two    & 32.0 & 94.0 & 90.0 & 81.3 & 94.0 \\
Hand over block     & 80.3 & 61.7 & 80.3 & 90.0 & 94.0 \\
Place dual shoes    & 72.0 & 70.0 & 76.0 & 76.7 & 76.0 \\
Place bread basket  & 50.7 & 67.3 & 64.3 & 71.0 & 64.7 \\
Move can pot        & 32.7 & 61.0 & 82.0 & 76.3 & 88.0 \\
Place fan           & 26.0 & 40.7 & 72.3 & 66.0 & 74.3 \\
Place phone stand   & 40.0 & 45.0 & 52.0 & 91.3 & 84.0 \\
\midrule
\textbf{Average}    & 43.5 & 64.9 & 74.1 & 76.1 & \textbf{79.0} \\
\bottomrule
\end{tabular*}
\end{table}

\subsubsection{LIBERO Experiments}
We evaluate our method on all four LIBERO task suites using a publicly available $\pi_{0.5}$ checkpoint trained on LIBERO. We compare adaptive execution against fixed execution chunks of 2, 4, 6, 8, and 10 actions without modifying the policy parameters. As shown in Tab.~\ref{tab:results_libero}, our method achieves the highest success rate on every suite and improves the overall average from 94.88\% for the strongest fixed-chunk baseline (chunk size 6) to 97.25\%. The improvement is most pronounced on LIBERO-Long, where our method reaches 94.5\%, exceeding the best fixed setting by 4.5 percentage points. These results indicate that a single static execution horizon cannot consistently accommodate tasks with different temporal structures, whereas adaptive action chunking transfers effectively to a separately trained policy and benchmark.

\begin{table}[t]
\centering
\caption{Success rate (\%) on the LIBERO benchmark.}
\label{tab:results_libero}
\small
\renewcommand{\arraystretch}{1.0}
\setlength{\tabcolsep}{2.5pt}
\begin{tabular*}{\linewidth}{@{\extracolsep{\fill}}lccccc}
\toprule
\textbf{Execution} & \textbf{Spatial} & \textbf{Object} & \textbf{Goal} & \textbf{Long} & \textbf{Avg.} \\
\midrule
Fixed-2  & 97.5 & 96.5 & 93.0 & 85.0 & 93.00 \\
Fixed-4  & 97.5 & 97.0 & 94.5 & 89.5 & 94.63 \\
Fixed-6  & 98.0 & 97.0 & 94.5 & 90.0 & 94.88 \\
Fixed-8  & 97.0 & 97.5 & 92.5 & 89.5 & 94.13 \\
Fixed-10 & 97.0 & 97.0 & 95.5 & 89.5 & 94.75 \\
\textbf{Ours} & \textbf{99.5} & \textbf{98.5} & \textbf{96.5} & \textbf{94.5} & \textbf{97.25} \\
\bottomrule
\end{tabular*}
\end{table}

\begin{table}[t]
\centering
    \small
    \caption{Success rate for real-world tasks.}
    \label{tab:results_tasks}
    \renewcommand{\arraystretch}{1.0}
    \setlength{\tabcolsep}{2.5pt} 
    \begin{tabular*}{\linewidth}{@{\extracolsep{\fill}}lcccccccc}
    \toprule
     & \multicolumn{8}{c}{\textbf{SR (\%) by Execution Chunk}} \\
    \cmidrule(lr){2-9}
    \textbf{Task Name} & \textbf{10} & \textbf{20} & \textbf{30} & \textbf{40} & \textbf{50} & \textbf{SA} & \textbf{MS} & \textbf{Ours} \\
    \midrule
    Hang Mug     & 55 & 45  & 60 & 60 & 65 & 60 & 20   & \textbf{75} \\
    Stack Cups   & 60 & 55 & 60 & 65 & 50 & 65 & 15  & \textbf{70} \\
    Tidy Pens    & 20  & \textbf{40}  & 35  & 15  & 10  & \textbf{40}  & 20  & \textbf{40}  \\
    Average       & 45 & 46.7 & 51.7 & 46.7 & 41.7 & 55 & 18.3 & \textbf{61.7} \\
    \bottomrule
    \end{tabular*}
\end{table}

\begin{table}[t]
    \centering
    \small
    \caption{Latency Analysis}
    \label{tab:latency_comparison}
    \renewcommand{\arraystretch}{1.0}
    \setlength{\tabcolsep}{4pt} 
    \begin{tabular*}{0.92\linewidth}{@{\extracolsep{\fill}}lcccc}
    \toprule
    Method & Fixed & MS & SA & \textbf{Ours} \\
    \midrule
    Latency / s $\downarrow$ & 0.2661 & 0.2956 & 0.2728 & \textbf{0.2725} \\
    \bottomrule
    \end{tabular*}
\end{table}

\subsection{Real-World Experiments}
\subsubsection{Platform and Tasks}
Our real-world system uses a dual-arm ARX R5 follower robot and an ARX X5 leader for bilateral teleoperation. Two wrist-mounted fisheye cameras provide local manipulation views, while a head-mounted RGB camera observes the full workspace. Images are captured at 20 Hz and the low-level control loop runs at 60 Hz. We evaluate three tasks shown in Fig.~\ref{fig:real_tasks_small}: \textit{Hang mug}, a single-arm task requiring precise placement on a mug-tree branch under partial occlusion; \textit{Tidy pens}, which transitions from single-arm placement to a dual-arm handover; and \textit{Stack cups}, a contact-rich dual-arm task requiring three cups to be stacked vertically.

\subsubsection{Training and Evaluation Protocol}
For each task, we collect 50 expert demonstrations and fine-tune a separate $\pi_{0.5}$ policy for 40,000 steps with batch size 32. Each execution method is evaluated over 20 independent trials per task.

As shown in Tab.~\ref{tab:results_tasks}, our method consistently improves success rates for both single-arm and dual-arm tasks. The average success rate across the three tasks increases by 10 percentage points compared to the best fixed-chunk configuration (chunk = 30) and by 6.7 percentage points relative to the strongest adaptive action chunking baseline (SA). It is also observed that the multi-sampling (MS) method struggles due to the fragmented execution resulting from the small chunk size. Using the configuration reported in the original literature, the robot exhibits intermittent pauses during movement, which impacts the overall success rate. These results demonstrate the effectiveness of utilizing the dispersion of cross-attention within the action expert as a metric to characterize the accuracy of predicted actions.

\subsection{Efficiency Analysis}
We analyze the computational latency for a single action chunk inference cycle on a single NVIDIA A100 GPU using $\pi_{0.5}$. As shown in Tab.~\ref{tab:latency_comparison}, the multi-sampling (MS) method increases the latency of each inference cycle by 29.5 ms. Although MS employs parallel execution for multiple sampling passes, such overhead remains considerable for high-frequency closed-loop control. Conversely, attention-based strategies, including the self-attention-based (SA) baseline and our approach, introduce less than 7 ms of additional latency relative to fixed chunking. By exploiting intrinsic model signals rather than executing redundant sampling passes, our method facilitates real-time dynamic action chunking with minimal computational cost.

\section{Conclusion}
This paper introduces an adaptive action chunking strategy using internal cross-attention dynamics in action expert modules. By calculating attention entropy from weight distributions, we characterize model predictive boundaries relative to observations. This enables a training-free dynamic truncation mechanism that adjusts the execution horizon during inference. Our experiments demonstrate improved success rates and robustness across multiple state-of-the-art VLA models in various multi-task settings. These improvements, achieved with negligible overhead, indicate that intrinsic epistemic uncertainty can be effectively used to enhance closed-loop robotic control.
\section{Limitations}
Our framework currently depends on identifiable cross-attention pathways. Architectures with highly interleaved or hierarchical structures present challenges in aggregating unified uncertainty signals. Furthermore, we have yet to evaluate our method on emerging World Action Models (WAMs)\cite{ye2026world,li2026causal,yuan2026fast} exhibiting distinct generative dynamics. Extending this adaptive strategy to such complex paradigms remains a promising future direction.



\bibliographystyle{IEEEtran}
\balance
\bibliography{references}
\end{document}